\documentclass[11pt]{article}

\usepackage{acl}
\usepackage{times}
\usepackage{latexsym}
\usepackage{xcolor}
\usepackage[T1]{fontenc}
\usepackage[utf8]{inputenc}
\usepackage{microtype}
\usepackage{amsmath}
\usepackage{amssymb}
\usepackage{booktabs}
\usepackage{array}
\usepackage{enumitem}
\usepackage{multirow}
\usepackage{tabularx}
\usepackage{graphicx}
\usepackage{dblfloatfix}
\usepackage{float}
\IfFileExists{fontawesome5.sty}{%
  \usepackage{fontawesome5}%
  \newcommand{\corrmark}{\textsuperscript{\scriptsize\faIcon{envelope}}}%
  \newcommand{\corrsymbol}{\faIcon{envelope}}%
}{%
  \newcommand{\corrmark}{\textsuperscript{\scriptsize @}}%
  \newcommand{\corrsymbol}{@}%
}
\usepackage{placeins}
\makeatletter
\renewcommand\section{\@startsection{section}{1}{\z@}{-1.6ex plus -0.4ex minus -.2ex}{1.0ex plus 0.2ex minus .2ex}{\large\bfseries\raggedright}}
\renewcommand\subsection{\@startsection{subsection}{2}{\z@}{-1.2ex plus -0.4ex minus -.2ex}{0.6ex plus .2ex}{\normalsize\bfseries\raggedright}}
\makeatother

\newcommand{\modelname}{Looped GPT-BERT}
\newcommand{\bertgpt}{\textsc{Bert}:\textsc{Gpt}}
\newcommand{\oldtext}[1]{}
\newcommand{\newtext}[1]{#1}

\newcolumntype{Y}{>{\centering\arraybackslash}X}

\title{Looped GPT-BERT: Trading Parameters for Computation in Small Language Modeling}

\author{
  Tingshuo Fan \quad
  Hongtao Mu \quad
  Tianyu Zhou \quad
  Hansen Liu \quad
  Tao Ji\corrmark \\
  College of Foreign Languages and Literature, Fudan University \\
  \small{
    \href{mailto:tsfan24@m.fudan.edu.cn}{tsfan24@m.fudan.edu.cn} \quad\quad
    \href{mailto:taoji@fudan.edu.cn}{taoji@fudan.edu.cn}
   }
}

\begin{document}
\maketitle

\begingroup
\renewcommand{\thefootnote}{\corrsymbol}
\footnotetext{Corresponding author.}
\endgroup

\setcounter{footnote}{0}

\begin{abstract}
\begingroup
\renewcommand{\newtext}[1]{}
\endgroup
\newtext{When training data are limited, increasing parameter count is not the only way to improve language-model performance. A small parameter set, when repeatedly applied, can also deliver comparable performance. We study Looped GPT-BERT in the BabyLM 2026 Strict-small setting, combining GPT-BERT's masked next-token and causal language-modeling objectives with depth-wise parameter sharing. We train on a preprocessed 7.48M-word English corpus and compare objective ratios, non-looped and looped architectures, and loop counts. Our final $4\times12$ model uses four physical layers for twelve recurrent traversals and contains 12.18M parameters. The BabyLM 2026 leaderboard reports an Overall Average of 35.42 and an NLP Average of 48.48. Compared with public BabyLM 10M Strict-small GPT-2 and GPT-BERT baselines, it achieves comparable performance on selected linguistic and downstream metrics, including BLiMP and GLUE, with fewer parameters. The loop ablations show that additional recurrent computation can improve training and preserve strong performance on selected linguistic tasks, whereas poorer performance on other tasks may reveal an inherent limitation of the looped design: using only a few physical layers restricts the model's representational space.}
\end{abstract}

\section{Introduction}

Language-model performance has commonly improved through jointly scaling model size, data, and training computation. Scaling-law studies describe predictable power-law relations between loss, parameter count, data, and compute \citep{kaplan2020scaling}, while compute-optimal analyses emphasize that parameter and token budgets should be allocated jointly \citep{hoffmann2022training}. These results largely concern much larger autoregressive models, where model capacity and per-example computation grow together. Under constrained data and parameter budgets, however, they need not: repeatedly applying the same parameters can create a deeper computational path without proportionally increasing the number of learned weights. We therefore ask whether recurrent computation induced by shared parameters can compensate for reduced physical depth and parameter capacity in low-resource language-model pretraining.

The BabyLM Challenge provides a controlled setting for this question by restricting pretraining text to a developmentally plausible scale and evaluating linguistic, downstream, and human-like behavior \citep{warstadt2023babylm}. We work in the BabyLM 2026 Strict-small track on a cleaned 7.48M-word English corpus. Our starting point is GPT-BERT \citep{charpentier2024gptbert}, which trains masked next-token prediction (MNTP) and causal language modeling (CLM) in one Transformer parameter stack. It consequently supports both bidirectional masked prediction and autoregressive scoring or generation. This hybrid objective gives a further dimension to study: when the backbone is looped, does the relative amount of masked and causal supervision change the capabilities the architecture acquires?

We introduce depth-wise recurrent parameter sharing on top of GPT-BERT. Universal Transformers repeatedly share transformations along depth \citep{dehghani2019universal}, and ALBERT shows that cross-layer sharing can substantially reduce encoder parameters \citep{lan2020albert}. Most directly, Looped Transformers inject a fixed input and the prior loop state into a shared backbone, increasing effective computational depth through iteration \citep{yang2024looped}. Their strongest evidence, however, comes from in-context algorithm-learning tasks such as linear functions, decision trees, and neural-network regression. Natural-language pretraining must simultaneously acquire lexical knowledge, syntactic regularities, world knowledge, and discourse state; whether those computations can be supported by repeated shared layers remains an open empirical question.

We address three questions: how the BERT:GPT training ratio changes the training objective and downstream abilities; whether effective depth formed by repeatedly applying a small number of physical layers can retain language ability with substantially fewer parameters; and whether training and downstream tasks continue to benefit as the number of loops increases. We summarize our contributions as follows:
\begin{itemize}[
    leftmargin=*,
    topsep=0pt,
    partopsep=0pt,
    parsep=0pt,
    itemsep=2pt
]
\setlength{\itemsep}{2pt}
\item We introduce Looped GPT-BERT, combining GPT-BERT's hybrid MNTP/CLM objective with depth-wise recurrent parameter sharing.
\item We study how the BERT:GPT objective ratio affects training and downstream behavior in the looped architecture.
\item We compare a 12-layer non-looped model with four physical layers applied 1/3/6/12 times, showing how recurrent computation closes the gap from fewer learned layers and where additional loops yield diminishing returns.
\item We evaluate the final model in the BabyLM Challenge, where its 12.18M parameters achieve comparable performance to public Strict-small baselines on linguistic and downstream tasks.
\end{itemize}

\begin{table*}[!t]
\centering
\small
\setlength{\tabcolsep}{4pt}
\begin{tabularx}{\textwidth}{>{\raggedright\arraybackslash}p{0.22\textwidth}XX}
\toprule
Cleaning operation & Before & After \\
\midrule
Style and grammar normalization & \textit{Where are you at?} & \textit{Where are you?} \\
Disfluency removal & \textit{a pound, or a hundred pounds today, is not the same} & \textit{A hundred pounds today is not worth the same amount later.} \\
Short-content filtering & \textit{Right.} & Removed when it lacks independent semantic content. \\
Case normalization & \textit{LET'S GIVE IT UP FOR DANNY TANNER} & \textit{Let's give it up for Danny Tanner.} \\
Structural-noise removal & \textit{=== Bullet the Blue Sky ===} & Header removed; article text retained. \\
Stage-direction removal & \textit{I'll be right back. [leaves room.]} & \textit{I'll be right back.} \\
Speaker normalization & \textit{*MOT: more what? *CHI: more tapioca.} & \textit{A: More what? B: More tapioca.} \\
Consecutive-turn merging & \textit{B: Yeah. B: I'm in Texas.} & \textit{B: Yeah, I'm in Texas.} \\
Duplicate removal & Repeated or near-identical sentences & One normalized instance. \\
\bottomrule
\end{tabularx}
\caption{Representative corpus-cleaning operations. Examples are shortened for presentation.}
\label{tab:cleaning}
\end{table*}

\section{Related Work}

\paragraph{Data-Efficient LM Pretraining}
BabyLM shifts attention from performance under web-scale training to learning efficiency under a controlled data budget. The first challenge showed that architecture, pretraining objective, preprocessing, and curriculum can all substantially affect small-model performance, with no single approach dominating every downstream task
\citep{warstadt2023babylm,hu-etal-2024-findings}. Subsequent work further found that the effect of data composition depends on model scale: a genre mixture suitable for one model size need not be optimal for another \citep{yam2024what}. Other BabyLM studies explore variation sets, child-inspired data and vocabulary choices, explicit linguistic information, and self-distillation as complementary routes to sample-efficient pretraining \citep{haga-etal-2024-babylm,ghanizadeh-dousti-2024-towards,edman-etal-2024-babylms,nair-etal-2024-babylm}. These approaches primarily improve sample efficiency through data selection, ordering, or objectives. Complementarily, we study parameter efficiency: keeping the training corpus and model width fixed, we replace part of the independently parameterized depth with repeated computation through shared physical layers.

\paragraph{Hybrid Causal and Masked Modeling}

CLM predicts the next token from left context, whereas conventional MLM predicts selected tokens at their own positions from bidirectional context. The two objectives therefore assign different meanings to the same output position in a decoder-style model. GPT-BERT resolves this mismatch with MNTP \citep{charpentier2024gptbert}: when token $x_{k+1}$ is masked, supervision is shifted left so that the hidden state at position $k$ predicts the original $x_{k+1}$. Consequently, causal and masked examples share a single next-token vocabulary projection and loss interface. GPT-mode rows retain their original tokens and use a lower-triangular attention mask; MNTP-mode rows corrupt selected input tokens, permit bidirectional attention, and compute loss only at the corresponding shifted next-token positions. Each training row is assigned one mode, rather than receiving both losses. AntLM similarly combines causal and masked language-modeling objectives in a BabyLM setting, alternating between them during training \citep{yu-etal-2024-antlm}.

\paragraph{Recurrent Depth and Parameter Sharing}
Universal Transformers update representations recurrently along depth while retaining parallel computation over sequence positions \citep{dehghani2019universal}; ALBERT uses cross-layer parameter sharing primarily for parameter efficiency and shows that the sharing strategy affects downstream behavior \citep{lan2020albert}. Looped Transformers use the recurrence $z_{t+1}=f(x+z_t)$, where $x$ is a fixed input embedding, $z_t$ is the loop state, and $f$ is a shared Transformer backbone \citep{yang2024looped}. This makes effective computational depth grow with loop count while the parameter count is governed mainly by the number of physical layers. \citet{kohli2026loop} provide a more direct natural-language reference: recurrent-depth Transformers can improve systematic generalization and depth extrapolation in implicit multi-hop reasoning, but excessive loops can also reduce prediction quality through overthinking. Neither line of work establishes that shared loops are uniformly suitable for lexical, syntactic, world-knowledge, and discourse-state learning. We therefore compare equal-application and equal-parameter configurations across linguistic and state-tracking tasks.

\section{Method}

\subsection{Corpus Preparation and Tokenization}

\oldtext{We use the English BabyLM Strict-small sources: BNC Spoken, CHILDES, Project Gutenberg, OpenSubtitles, Simple English Wikipedia, and Switchboard. The team's preprocessing pipeline combines rule-based filtering with rewriting-based cleaning to remove format noise while retaining source semantics and discourse structure across conversational transcripts, subtitles, encyclopedia text, and books.} \newtext{We construct our training corpus by selecting and cleaning six source corpora: BNC Spoken, CHILDES, Project Gutenberg, OpenSubtitles, Simple English Wikipedia, and Switchboard. The resulting \texttt{new\_data1} corpus is produced by our rule-based normalization and filtering pipeline, which removes redundancy, short or low-quality content, case and structural noise, and normalizes dialogue formatting. The preprocessing code and source files are available in our \href{https://github.com/JT-Ushio/babylm26-nlp-spring}{preprocessing repository}.} \textbf{Table~\ref{tab:cleaning}} summarizes the operations and representative input--output examples. \newtext{After cleaning, the six selected corpora yield 7,482,189 whitespace-delimited words. We use this complete post-cleaning corpus directly for pretraining rather than adding or padding data to reach the 10M-word upper bound. Their source-wise counts are reported in} \textbf{Table~\ref{tab:data}}.

\oldtext{We use two byte-pair encoding (BPE) tokenizers. The official 16k GPT-BERT tokenizer is used only for early non-looped ratio experiments. The 8k tokenizer is trained on the \texttt{new\_data1} corpus and is used for all main and loop experiments. This choice balances subword statistics and parameter allocation under the 7.48M-word data budget: an overly large vocabulary creates many low-frequency subwords with limited learning signal and allocates a substantial fraction of a small model to embeddings. At hidden size 384, 8,192 tokens correspond to approximately 3.15M tied input/output embedding parameters, about one quarter of the final 12.18M-parameter model. The 8k setting reduces this embedding cost without making the 128-token context window cover substantially fewer words. We report 16k and 8k results separately and use them only to assess whether objective-ratio trends are consistent, not to claim a tokenizer advantage.}
\newtext{We use two byte-pair encoding (BPE) tokenizers. The official 16k GPT-BERT tokenizer is used only for early non-looped ratio experiments, while the 8k tokenizer trained on the \texttt{new\_data1} corpus is used for the main and loop experiments. BPE provides a subword representation that can handle words beyond a fixed vocabulary \citep{sennrich-etal-2016-neural}. We choose the 8k vocabulary as a parameter-allocation trade-off for this small-data setting: with hidden size 384, its tied embedding table contains $8{,}192\times384\approx3.15$M parameters. In our corpus, reducing the vocabulary from 16k to 8k increases the measured fertility from 1.4381 to 1.4787 tokens per whitespace-delimited word, a 2.82\% increase. This indicates a modest change in word-level context coverage for the fixed 128-token input, while avoiding a larger embedding allocation.}

\begin{table}[H]
\centering
\small
\begin{tabular}{lr}
\toprule
Source & Words \\
\midrule
BNC Spoken & 619,847 \\
CHILDES & 1,779,027 \\
Gutenberg & 2,502,709 \\
OpenSubtitles & 1,174,391 \\
Simple Wikipedia & 1,386,065 \\
Switchboard & 20,150 \\
\midrule
Total & 7,482,189 \\
\bottomrule
\end{tabular}
\caption{Word counts after preprocessing.}
\label{tab:data}
\end{table}

\subsection{Hybrid GPT-BERT Objective}

Let $x=(x_1,\ldots,x_T)$ be a token sequence and $\theta$ the shared model parameters. For GPT-mode rows, a causal attention mask is used and every non-padding next token is supervised:
\begin{equation}
\mathcal{L}_{\mathrm{GPT}}
=-\sum_{t=1}^{T-1}\log p_\theta(x_{t+1}\mid x_{\leq t}).
\end{equation}

For BERT-mode rows, predictable positions are sampled with a masking probability that decreases linearly from 0.3 to 0.15. Of the selected tokens, 80\% are replaced by the mask token, 10\% by a random token, and 10\% are left unchanged. Bidirectional attention is permitted. The target is shifted to the preceding hidden-state position so that it remains aligned with the decoder-style next-token head. If $\tilde{x}$ denotes the corrupted input, the objective is
\begin{equation}
\mathcal{L}_{\mathrm{BERT}}
=-\sum_{t\in M}\log p_\theta(x_t\mid \tilde{x}_{\setminus t}).
\end{equation}
Only shifted positions corresponding to selected masks contribute to this loss; all other labels are set to the ignore index. Thus, unmasked BERT positions never enter the loss. A batch-level ratio of $r_{\mathrm{B}}:r_{\mathrm{G}}$ assigns complete rows to the two modes, and the optimized objective averages valid supervised tokens with a $z$-loss regularizer weighted by $10^{-4}$. Our final setting uses $1{:}3$, so one quarter of rows use MNTP and three quarters use CLM.

\subsection{Looped Transformer Backbone}

The non-looped baseline contains twelve independent Transformer layers. The looped backbone instead contains four independently parameterized physical layers, each with attention and feed-forward sublayers. Each loop adds the static token representation $x$ to the preceding loop state $z_t$ and applies the same four-layer stack $F_\theta$:
\begin{equation}
\label{eq:loop}
\begin{aligned}
z_0 &= 0,\\
z_{t+1} &= F_\theta(x+z_t),
\quad t=0,\ldots,L-1.
\end{aligned}
\end{equation}
Thus, $z_{t+1}$ is the final representation produced by the current traversal and is simply carried into the next pass; it is not an additional network operation. Within physical layer $i$, attention and feed-forward transformations are applied sequentially:
\begin{equation}
\label{eq:physical}
\begin{aligned}
u_t^{(i)} &= \operatorname{DWA}_{i,A}\!\left(h_t^{(i)}+
\operatorname{Attn}_i(h_t^{(i)})\right),\\
h_t^{(i+1)} &= \operatorname{DWA}_{i,F}\!\left(u_t^{(i)}+
\operatorname{FFN}_i(u_t^{(i)})\right).
\end{aligned}
\end{equation}
GPT-BERT's dynamic weighted accumulation (DWA) is a learnable short-range residual mixer, not an additional recurrent state. We restrict DWA to representation fusion within each physical layer and reinitialize it on every loop; information across loops is carried exclusively by $z_t$. A $4\times L$ model therefore has four sets of Transformer parameters but performs $4L$ layer applications. The $4\times3$, $4\times6$, and $4\times12$ variants perform 12, 24, and 48 layer applications, respectively, while keeping the same parameter count. This ``effective depth'' counts layer applications only: repeatedly applying shared parameters is not equivalent to a non-shared Transformer with the same number of layers.

\subsection{Training Configuration}

\textbf{Table~\ref{tab:hyperparameters}} lists the main configuration. Beyond the structural hyperparameters shown there, optimization uses LAMB \citep{you2020large}, a peak learning rate of 0.0141, a minimum learning rate of 0.00141, and cosine decay for 2,600 steps followed by a constant minimum rate. We select the peak learning rate and decay horizon through the controlled studies in Section~\ref{sec:optimization}. \newtext{All pretraining and fine-tuning runs use random seed 42.}

\begin{table}[H]
\centering
\small
\begin{tabular}{lr}
\toprule
Hyperparameter & Value \\
\midrule
Physical layers & 4 \\
Loop iterations & 12 \\
Hidden size & 384 \\
FFN size & 1,280 \\
Attention heads & 6 \\
Vocabulary & 8,192 \\
Maximum sequence length & 128 \\
Batch size per GPU & 32 \\
Number of GPUs & 8 \\
Global batch (tokens) & 32,768 \\
Epochs & 10 \\
Optimizer & LAMB \\
Peak learning rate & 0.0141 \\
Decay steps & 2,600 \\
Minimum learning rate & 0.00141 \\
Gradient clipping & 2.0 \\
Parameters & 12.18M \\
\bottomrule
\end{tabular}
\caption{Main pretraining configuration.}
\label{tab:hyperparameters}
\end{table}

\section{Experiments}

\subsection{Evaluation Protocol}

We follow the BabyLM 2026 Strict-small evaluation pipeline. The final checkpoint receives the full evaluation: the zero-shot suite includes BLiMP and BLiMP Supplement minimal-pair syntax judgments \citep{warstadt2020blimp}, EWoK world knowledge \citep{ivanova2024ewok}, Entity Tracking, COMPS conceptual properties, GlobalPIQA commonsense reasoning, Reading, and Age of Acquisition (AoA). Fine-tuning covers BoolQ, MNLI, MRPC, MultiRC, QQP, RTE, and WSC, the GLUE/SuperGLUE-style natural-language-understanding subset used by the challenge \citep{wang2019glue}. We report the leaderboard aggregates, where NLP Average pools NLP tasks and Human-like Average includes Reading, AoA, and related measures. We use the official causal backend for all reported GPT-BERT results.

\oldtext{In addition to the final \emph{main} revision, we save \emph{chck\_1M} through \emph{chck\_10M} at one-million-word intervals and \emph{chck\_20M} through \emph{chck\_70M} at ten-million-word intervals, and run the official fast evaluation on these checkpoints.} \newtext{Beyond the final \emph{main} revision, we retain intermediate checkpoints at one-million-word intervals from \emph{chck\_1M} through \emph{chck\_10M}, and at ten-million-word intervals from \emph{chck\_20M} through \emph{chck\_70M}. We run the official fast evaluation on each of these checkpoints.} The ten-epoch training run contains approximately 74.8M word presentations, so \emph{chck\_70M} is the last complete ten-million-word milestone. The final model and all intermediate revisions are released in one public Hugging Face repository.

\subsection{BabyLM 2026 Leaderboard Results}

\begin{table*}[t]
\centering
\scriptsize
\setlength{\tabcolsep}{1pt}
\resizebox{\textwidth}{!}{%
\begin{tabular}{lrr*{9}{c}}
\toprule
Model & Params & Context & Overall & NLP & BLiMP & Sup. & EWoK & Entity & COMPS & GPIQA & GLUE \\
\midrule
\newtext{Official GPT-2 baseline} & \newtext{124M} & \newtext{512} & \newtext{37.38} & \newtext{48.99} & \newtext{65.23} & \newtext{57.25} & \newtext{50.63} & \newtext{19.10} & \newtext{51.81} & \newtext{35.09} & \newtext{63.80} \\
\newtext{Official GPT-BERT causal-focus} & \newtext{31M} & \newtext{128$\rightarrow$512} & \newtext{28.46} & \newtext{35.64} & \newtext{71.66} & \newtext{63.21} & \newtext{49.49} & \newtext{--} & \newtext{--} & \newtext{--} & \newtext{65.13} \\
\newtext{Looped GPT-BERT (ours)} & \newtext{\textbf{12.18M}} & \newtext{\textbf{128}} & \newtext{35.42} & \newtext{48.48} & \newtext{71.19} & \newtext{55.06} & \newtext{49.10} & \newtext{15.78} & \newtext{51.01} & \newtext{34.67} & \newtext{62.55} \\
\bottomrule
\end{tabular}%
}
\caption{BabyLM 2026 Strict-small leaderboard comparison on aggregate and selected task-level metrics. ``Context'' denotes the maximum training sequence length, ``Sup.'' denotes BLiMP Supplement, ``Entity'' denotes Entity Tracking, ``GPIQA'' denotes GlobalPIQA, and ``GLUE'' denotes the official (Super)GLUE aggregate. Dashes indicate results not reported for the corresponding public entry. Aggregate scores for entries with unreported tasks are not directly comparable.}
\label{tab:leaderboard}
\end{table*}

\textbf{Table~\ref{tab:leaderboard}} reports the final $4\times12$ Looped GPT-BERT alongside two publicly released BabyLM Strict-small references: the official GPT-2 baseline and the GPT-BERT causal-focus baseline. The leaderboard snapshot was accessed on July 21, 2026.\footnote{\url{https://huggingface.co/spaces/BabyLM-community/BabyLM-Leaderboard-2026}}

\oldtext{With 12.18M parameters, \modelname{} uses 12.4\% of the 98M GPT-2 baseline's parameters while retaining 94.8\% of its Overall Average (35.42 versus 37.38) and 99.0\% of its NLP Average (48.48 versus 48.99); it also has the highest BLiMP score in the table. Relative to the 31M GPT-BERT mixed causal baseline, it reduces parameter count by 60.7\% and raises the reported Overall and NLP averages by 7.25 and 13.18 points.} \newtext{With 12.18M parameters, the final model is substantially smaller than both public references. It reaches 71.19 on BLiMP, compared with 65.23 for GPT-2 and 71.66 for GPT-BERT causal-focus, while its GLUE aggregate is 62.55, compared with 63.80 and 65.13. This pattern illustrates the trade-off studied here: at the cost of additional computation, repeated application of shared layers can retain strong performance on selected tasks with a smaller independently parameterized model, with gains varying across tasks.}

\subsection{Effect of the BERT:GPT Ratio}

We examine the BERT:GPT ratio under two tokenizers. The official 16k tokenizer is used for early non-looped ratio exploration, whereas the 8k tokenizer trained on the \texttt{new\_data1} corpus is used for all main and loop experiments. The $15{:}1$ configuration is BERT-heavy, $1{:}1$ is balanced, and $1{:}3$ is GPT-heavy. The two settings test whether the ratio trend is consistent; they are not used to compare tokenizer quality. \textbf{Tables~\ref{tab:ratio8k-train} and~\ref{tab:ratio8k-zero}} report the principal 8k comparison, including training endpoints and available zero-shot results.

\begin{table}[!htbp]
\centering
\small
\begin{tabular}{lrrr}
\toprule
\bertgpt{} & Params & Loss & Acc. \\
\midrule
1:1 & 29.9M & 3.9925 & 30.03 \\
1:3 & 29.9M & \textbf{3.3625} & \textbf{36.47} \\
\bottomrule
\end{tabular}
\caption{Non-looped training results across objective ratios with the custom 8k tokenizer.}
\label{tab:ratio8k-train}
\end{table}

\begin{table}[!htbp]
\centering
\small
\begin{tabular}{lrrrr}
\toprule
\bertgpt{} & BLiMP & Sup. & COMPS & Entity \\
\midrule
1:1 & 63.77 & 55.06 & 50.87 & \textbf{38.66} \\
1:3 & \textbf{70.54} & \textbf{57.19} & \textbf{51.41} & 22.87 \\
\bottomrule
\end{tabular}
\caption{Non-looped zero-shot results across objective ratios with the custom 8k tokenizer.}
\label{tab:ratio8k-zero}
\end{table}

For context, \textbf{Table~\ref{tab:ratio16k}} reports the corresponding early experiments with the official 16k tokenizer. Increasing the proportion of GPT rows lowers the final mixed training loss in both tokenizer settings, with $1{:}3$ giving the strongest endpoint.

\begin{table}[!htbp]
\centering
\small
\begin{tabular}{lrr}
\toprule
\bertgpt{} & Final loss & Token accuracy \\
\midrule
15:1 & 4.0039 & 33.42 \\
1:1 & 3.9711 & 31.50 \\
1:3 & \textbf{3.7925} & 32.37 \\
\bottomrule
\end{tabular}
\caption{Non-looped training results across objective ratios with the official 16k tokenizer.}
\label{tab:ratio16k}
\end{table}

The $1{:}3$ ratio improves BLiMP, BLiMP Supplement, and COMPS under the 8k tokenizer, but reduces Entity Tracking from 38.66 to 22.87. This pattern suggests that objective mixing changes the distribution of learned capabilities rather than merely the aggregate loss. More CLM supervision repeatedly trains left-to-right next-token prediction and is therefore closely aligned with sequential generation, local dependencies, and causal syntactic scoring. More MNTP supervision requires reconstructing a target from both sides of its context and may better support integrating multiple positions and relations, which is useful for tracking entities and their states. We therefore use $1{:}3$ as the main configuration for subsequent loop experiments targeting generative and syntactic ability.

\subsection{Effect of Looped Depth}
\label{sec:loops}

\begin{figure*}[t]
\centering
\includegraphics[width=0.96\textwidth]{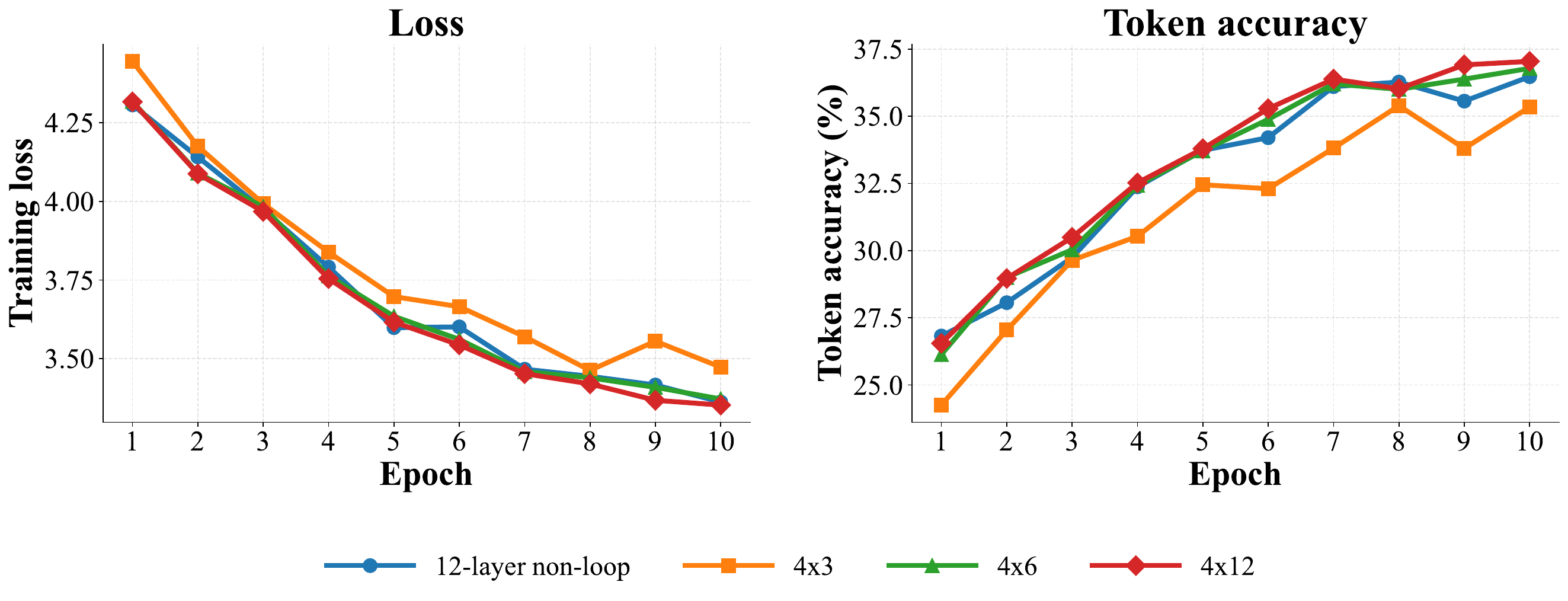}
\caption{Training loss and token accuracy for the 8k-tokenizer, $1{:}3$ objective across loop depths. The looped models use four physical layers; the non-looped model uses twelve independently parameterized layers.}
\label{fig:looptrain}
\end{figure*}

\newtext{We first examine training loss and token accuracy under the $1{:}3$ objective. \textbf{Figure~\ref{fig:looptrain}} plots how these two measures change over training for the 12-layer non-looped model and for four-layer models with $1$, $3$, $6$, or $12$ applications per forward pass. The $4\times3$ and 12-layer non-looped models both execute twelve layer applications per forward pass, while the looped model uses only four independently parameterized layers.}

\oldtext{The $4\times3$ and 12-layer non-looped models both execute twelve layer applications per forward pass, but the former retains only four physical layers; $4\times6$ and $4\times12$ increase computation further without increasing the number of learned physical-layer parameters. The training curves in Figure~\ref{fig:looptrain} show the role of recurrent depth more clearly. The $4\times1$ control, which uses the same four physical layers without repeated application, ends with a higher loss and lower token accuracy than the deeper configurations. Reusing the physical layers three times closes much of this gap, while the changes from $4\times3$ to $4\times6$ and from $4\times6$ to $4\times12$ are progressively smaller. Thus, recurrent computation recovers much of the training behavior of the non-looped model, but with diminishing returns as loop depth increases.}
\newtext{The $4\times1$ control is visibly less stable at the beginning of training: it maintains a higher loss and lower token accuracy than the models with repeated layer applications. Increasing the loop count lowers the training loss and generally raises token accuracy, bringing the looped models closer to the 12-layer non-looped trajectory. The largest improvement occurs when moving from $4\times1$ to $4\times3$; the changes from $4\times3$ to $4\times6$ and from $4\times6$ to $4\times12$ are smaller. Thus, repeated computation recovers much of the training behavior of the non-looped model, while additional loops provide diminishing returns.}

\begin{table}[!t]
\centering
\scriptsize
\setlength{\tabcolsep}{3pt}
\begin{tabular}{lrrrrr}
\toprule
\multicolumn{6}{c}{\bertgpt{} = 1:3} \\
\midrule
Metric & Non-loop & $4\times1$ & $4\times3$ & $4\times6$ & $4\times12$ \\
\midrule
BLiMP & 70.54 & \newtext{65.08} & \textbf{\newtext{71.21}} & 70.49 & 71.19 \\
BLiMP Sup. & \textbf{57.19} & \newtext{55.90} & \newtext{55.73} & 55.37 & 55.06 \\
COMPS & 51.41 & \newtext{51.21} & \newtext{51.40} & \textbf{51.69} & 51.01 \\
Entity Tracking & 22.87 & \newtext{17.38} & \newtext{16.47} & \textbf{28.66} & 15.78 \\
Eye Tracking & 8.61 & \textbf{\newtext{8.62}} & \newtext{7.85} & \textbf{8.62} & 8.25 \\
Self-paced Reading & 4.03 & \newtext{3.88} & \newtext{3.68} & 4.16 & \textbf{4.24} \\
\midrule
\multicolumn{6}{c}{\bertgpt{} = 1:1} \\
\midrule
Metric & Non-loop & $4\times1$ & $4\times3$ & $4\times6$ & $4\times12$ \\
\midrule
BLiMP & 63.77 & \newtext{62.28} & \newtext{64.45} & \textbf{69.38} & \newtext{64.21} \\
BLiMP Sup. & 55.06 & \textbf{\newtext{57.03}} & \newtext{54.96} & 56.11 & \newtext{56.07} \\
COMPS & 50.87 & \newtext{50.44} & \newtext{50.95} & \textbf{51.42} & \newtext{50.03} \\
Entity Tracking & \textbf{38.66} & \newtext{17.52} & \newtext{16.60} & 13.10 & \newtext{17.15} \\
Eye Tracking & 7.86 & \newtext{7.64} & \textbf{\newtext{7.88}} & 7.62 & \newtext{7.55} \\
Self-paced Reading & 3.80 & \textbf{\newtext{3.84}} & \newtext{3.57} & 3.56 & \newtext{3.63} \\
\bottomrule
\end{tabular}
\caption{Zero-shot comparison across objective ratios and loop counts.}
\label{tab:loopzero}
\end{table}
\newtext{We next turn to the zero-shot task results in \textbf{Table~\ref{tab:loopzero}}. The table compares the 12-layer non-looped model with four physical layers applied $1$, $3$, $6$, or $12$ times, using BERT:GPT ratios of $1{:}3$ in the upper block and $1{:}1$ in the lower block. Under $1{:}3$, increasing the loop count from $4\times1$ to $4\times3$ substantially improves BLiMP, from 65.08 to 71.21, slightly exceeding the non-looped score of 70.54. Increasing the depth beyond $4\times3$ does not produce a consistent additional gain: BLiMP is 70.49 for $4\times6$ and 71.19 for $4\times12$. Under $1{:}1$, $4\times6$ gives the highest BLiMP score (69.38), compared with 63.77 for the non-looped model and 62.28--64.45 for the other looped settings. Overall, the looped models preserve or improve several syntactic and reading-related scores despite using fewer independently parameterized layers, while the effect of recurrent depth varies across tasks.}

\newtext{Unlike the syntactic measures above, Entity Tracking is more sensitive to parameter sharing. Overall, the looped variants are weaker than the corresponding non-looped model on this task, with one exception: under $1{:}3$, $4\times6$ reaches 28.66, compared with 22.87 for the non-looped model. Even in this setting, the score falls to 15.78 with $4\times12$, so the additional computation from six to twelve applications does not preserve the $4\times6$ improvement. Under $1{:}1$, all looped variants remain below the non-looped score of 38.66, with scores between 13.10 and 17.52. One possible explanation is that entity tracking requires distinguishing and updating multiple entities, their states, and state changes \citep{kim2023entity}. Reusing four attention/FFN parameter sets rather than twelve independent sets may leave less representational space for the layer-specific transformations needed to maintain multiple entities and relations. This contrast suggests that recurrent depth can compensate for reduced parameterization on some linguistic tasks, but cannot uniformly replace the representational flexibility of independently parameterized layers. Considering both downstream performance and inference cost, $4\times6$ is a practical intermediate configuration, but the table also shows that the preferred loop depth depends on the objective ratio and task.}

\subsection{Inference Cost}

\newtext{We measure inference cost for the 12-layer non-looped model and the $4\times3$, $4\times6$, and $4\times12$ looped models with the same batch size of 32, sequence length of 128, 20 warmup steps, and 100 timed forward passes on one RTX 3090. As shown in \textbf{Table~\ref{tab:inference_speed}}, $4\times3$ is better than the 12-layer non-looped model on every measured systems metric: it has lower latency, higher token throughput, and lower peak memory use. Increasing the loop count from $4\times3$ to $4\times6$ and $4\times12$ roughly doubles the latency at each step, while throughput decreases and memory use increases. Considering both downstream performance and inference speed, $4\times6$ provides a practical compromise: it is slower than $4\times3$ but performs better on some downstream tasks, while remaining substantially faster than $4\times12$.}

\begin{table}[t]
\centering
\scriptsize
\setlength{\tabcolsep}{2pt}
\begin{tabular}{lrrrr}
\toprule
Model & Latency (ms) & Tokens/s & Memory (GiB) \\
\midrule
\newtext{12-layer non-loop} & \newtext{58} & \newtext{70,571} & \newtext{0.68} \\
\newtext{$4\times3$} & \newtext{\textbf{55}} & \newtext{\textbf{74,128}} & \newtext{\textbf{0.51}} \\
\newtext{$4\times6$} & \newtext{108} & \newtext{37,966} & \newtext{0.72} \\
\newtext{$4\times12$} & \newtext{212} & \newtext{19,295} & \newtext{1.15} \\
\bottomrule
\end{tabular}
\caption{Inference speed measured with batch size 32, sequence length 128, 20 warmup steps, and 100 timed forward passes on one RTX 3090.}
\label{tab:inference_speed}
\end{table}

\section{Optimization Ablations}
\label{sec:optimization}

\subsection{Selecting the Peak Learning Rate}

We first select a peak learning rate that lowers the training objective quickly without producing clear instability. Before full training, we run four-epoch non-looped sweeps at 0.0075, \oldtext{0.0100}\newtext{0.01}, 0.0141, and \oldtext{0.0200}\newtext{0.02} with all other settings fixed. As \textbf{Table~\ref{tab:lrsweep}} shows, the first three settings decrease stably, whereas \oldtext{0.0200}\newtext{0.02} degrades after approximately step 600 and produces gradient-norm spikes. Although \oldtext{0.0100}\newtext{0.01} has the highest final token accuracy, 0.0141 achieves the lowest final loss (3.6847) and is selected as the peak learning rate for the main experiments.
\newtext{This sweep provides a preliminary learning-rate choice based on the non-looped model and does not fully optimize the rate for looped models. We examine the effect of decay speed on training stability separately below.}

\begin{table}[!htbp]
\centering
\small
\begin{tabular}{rrr}
\toprule
Peak LR & Final loss & Final accuracy \\
\midrule
0.0075 & 3.7442 & 32.66 \\
0.0100 & 3.6931 & \textbf{33.07} \\
0.0141 & \textbf{3.6847} & 32.95 \\
0.0200 & 4.5864 & 22.84 \\
\bottomrule
\end{tabular}
\caption{Four-epoch peak learning-rate sweep.}
\label{tab:lrsweep}
\end{table}

\subsection{Learning-Rate Decay and Stability}

\newtext{We select two candidate decay lengths from the training budget: 4,080 steps spans the full optimizer-step budget of the run, whereas 2,600 steps reaches the minimum earlier and remains there for the rest of training.} With total training steps fixed, \emph{lr\_schedule\_steps} determines how quickly cosine decay reaches its minimum: a smaller value decays earlier and remains at the minimum longer, while a larger value retains a higher learning rate later in training. We compare these decay rates in $4\times6$ and $4\times12$ models. The main 2,600-step schedule reaches 10\% of the peak learning rate and then holds 0.00141; the slower 4,080-step schedule maintains a higher learning rate into the later stages. Because shared physical layers are repeatedly applied across loops, the effect of one parameter update can be propagated through multiple layer applications. A high early learning rate may therefore amplify update oscillations, while earlier decay may help shared parameters enter a useful region more smoothly.

\begin{figure*}[t]
\centering
\includegraphics[width=0.96\textwidth]{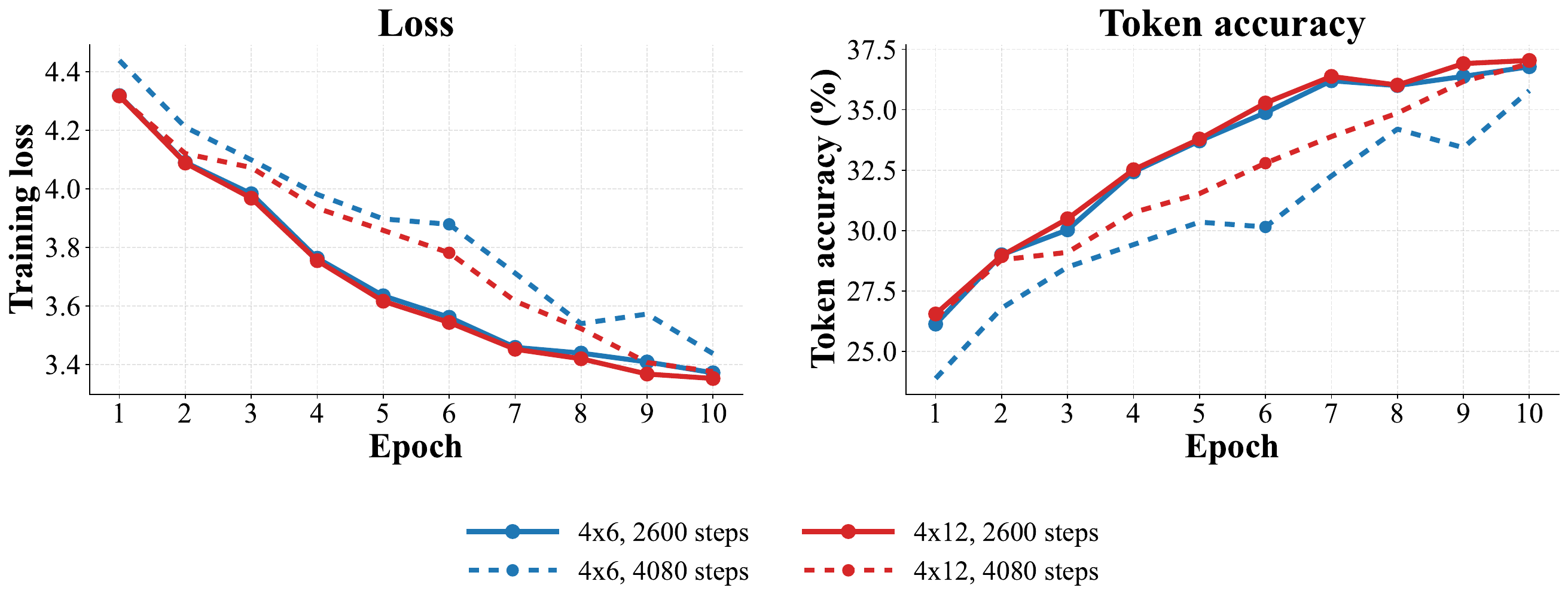}
\caption{Training loss and token accuracy for the learning-rate decay ablation. Solid and dashed lines denote 2,600-step and 4,080-step decay, respectively; colors distinguish $4\times6$ and $4\times12$.}
\label{fig:decay}
\end{figure*}

\textbf{Figure~\ref{fig:decay}} shows that the 2,600-step schedule has its clearest advantage in early and middle training: loss declines faster and token accuracy rises sooner. Earlier decay, however, can reduce the opportunity to explore alternative parameter regions and leave the model near a suboptimal solution. The 4,080-step schedule gradually catches up later, consistent with its higher late-stage learning rate preserving additional exploration. Nevertheless, the 2,600-step curves are smoother overall and show no clear final-metric disadvantage. We therefore choose 2,600 steps as a practical compromise between stability and late-stage exploration under the current compute budget and single-seed setting, not as a schedule that is uniformly superior at every point in training.

\subsection{Qualitative Generation Examples}

The causal interface also supports open-ended completion. \textbf{Table~\ref{tab:generation}} presents selected completions from the final $1{:}3$, $4\times12$ model using temperature 0.9, top-$k$ 40, top-$p$ 0.95, and seed 42. The examples illustrate locally coherent continuation and appropriate termination in short responses; they are qualitative examples rather than a controlled generation evaluation.

\begin{table}[!htbp]
\centering
\small
\begin{tabularx}{\columnwidth}{>{\raggedright\arraybackslash}p{0.38\columnwidth}X}
\toprule
Prompt & Sampled completion \\
\midrule
\textit{I want to eat some} & \textit{dinner, all right?} \\
\textit{After the rain stopped, the children} & \textit{began to run for a walk.} \\
\textit{She put the toy in the} & \textit{drawer, and gave the doll a glass of milk.} \\
\textit{The bird flew over the} & \textit{mountains and in the morning the rain came in.} \\
\bottomrule
\end{tabularx}
\caption{Selected sampled completions from the final $1{:}3$, $4\times12$ model.}
\label{tab:generation}
\end{table}

\section{Conclusion}

We combine GPT-BERT's hybrid CLM/MNTP training with depth-wise recurrent parameter sharing to study a ``fewer parameters, more computation'' language-model design under BabyLM Strict-small constraints. Reusing four physical layers for twelve loops yields a 12.18M-parameter model whose training objective is close to that of a 29.9M non-looped model. The final model reaches 71.19 on BLiMP and 62.55 on GLUE, compared with 71.66 and 65.13 for the official GPT-BERT causal-focus reference, despite its smaller parameter budget. On the reported task-level metrics, looped models preserve or improve selected syntactic abilities, while Entity Tracking becomes weaker as a result of parameter sharing. The ablations further show that gains from additional loops saturate quickly and differ across tasks. Looped GPT-BERT is therefore an architecture with clear advantages and limitations: under constrained data and parameter budgets, it can support syntactic understanding and generation, but its objective ratio and recurrent depth require further adjustment.

\section*{Limitations}

\begin{itemize}[
    leftmargin=*,
    topsep=0pt,
    partopsep=0pt,
    parsep=0pt,
    itemsep=2pt
]
    \item \textbf{Data and tokenizer.} The main experiments use one cleaned corpus and one 8k tokenizer. We do not independently ablate raw data, rule-based cleaning, alternative cleaning procedures, and tokenization.
    \item \textbf{Training variance.} Each model is trained with a single random seed, so sub-percentage-point differences may fall within training variance.
    \item \textbf{Compute matching.} The $4\times12$ model performs substantially more layer applications and computation than the 12-layer non-looped baseline. This is a parameter-efficiency study, not a FLOP-matched comparison.
    \item \textbf{Objective mixing.} Changing the \bertgpt{} ratio also changes supervision density and attention visibility; mixed training losses are therefore not perfectly homogeneous across ratios.
    \item \textbf{Capability boundary.} The final model is weak on AoA and Entity Tracking, indicating limited cognitive-similarity and state-tracking ability.
    \item \textbf{Scope.} Our conclusions are limited to approximately 12M parameters, a 7.48M-word corpus, ten epochs, and a maximum sequence length of 128; they should not be directly extrapolated to larger models or longer contexts.
\end{itemize}

\section*{Ethics Statement}

The model uses English text supplied by, or derived from, the BabyLM Challenge and introduces no newly collected personal data. The sources include books, subtitles, encyclopedic text, and dialogue, so they may retain social biases or inappropriate content from the original material; automated cleaning cannot guarantee their removal. Cleaning may also remove dialectal, conversational, or minority-language patterns and introduce stylistic preferences through normalization. This small model is intended for research and should not be deployed directly in high-stakes applications.

\section*{Reproducibility Statement}

The Hugging Face repository contains the final model and sixteen intermediate checkpoints. The \emph{main} revision stores the final model; \emph{chck\_1M}--\emph{chck\_10M} are one-million-word checkpoints, and \emph{chck\_20M}--\emph{chck\_70M} are ten-million-word checkpoints. It also includes \emph{configuration\_gpt\_bert.py}, \emph{modeling\_gpt\_bert.py}, and tokenizer files, allowing the model to be loaded with \emph{trust\_remote\_code=True}.\footnote{\url{https://huggingface.co/Survivor613/BabyLM2026-Strict-Small-looped_GPT-BERT}} Training configurations and scripts are available in the accompanying code repository,\footnote{\url{https://github.com/Survivor613/BabyLM-2026}} and the data-cleaning workflow is documented separately.\footnote{\url{https://github.com/JT-Ushio/babylm26-nlp-spring}}

\section*{Acknowledgments}
The authors thank the reviewers for their helpful comments and suggestions, as well as the BabyLM organizers for maintaining the datasets, evaluation pipeline, and leaderboard. This work was partially funded by the National Natural Science Foundation of China (No. 62506079).

\bibliography{custom}

@inproceedings{sennrich-etal-2016-neural,
  title = {Neural Machine Translation of Rare Words with Subword Units},
  author = {Sennrich, Rico and Haddow, Barry and Birch, Alexandra},
  booktitle = {Proceedings of the 54th Annual Meeting of the Association for Computational Linguistics (Volume 1: Long Papers)},
  pages = {1715--1725},
  year = {2016},
  publisher = {Association for Computational Linguistics},
  doi = {10.18653/v1/P16-1162},
  url = {https://aclanthology.org/P16-1162/}
}

@article{kaplan2020scaling,
  title = {Scaling Laws for Neural Language Models},
  author = {Kaplan, Jared and McCandlish, Sam and Henighan, Tom and Brown, Tom B. and Chess, Benjamin and Child, Rewon and Gray, Scott and Radford, Alec and Wu, Jeffrey and Amodei, Dario},
  journal = {arXiv preprint arXiv:2001.08361},
  year = {2020},
  url = {https://arxiv.org/abs/2001.08361}
}

@inproceedings{hoffmann2022training,
  title = {Training Compute-Optimal Large Language Models},
  author = {Hoffmann, Jordan and Borgeaud, Sebastian and Mensch, Arthur and Buchatskaya, Elena and Cai, Trevor and Rutherford, Eliza and de Las Casas, Diego and Hendricks, Lisa Anne and Welbl, Johannes and Clark, Aidan and others},
  booktitle = {Advances in Neural Information Processing Systems},
  volume = {35},
  pages = {30016--30030},
  year = {2022}
}

@inproceedings{warstadt2023babylm,
  title = {Findings of the {B}aby{LM} Challenge: Sample-Efficient Pretraining on Developmentally Plausible Corpora},
  author = {Warstadt, Alex and Mueller, Aaron and Choshen, Leshem and Wilcox, Ethan and Zhuang, Chengxu and Ciro, Juan and Mosquera, Rafael and Paranjape, Bhargavi and Williams, Adina and Linzen, Tal and Cotterell, Ryan},
  booktitle = {Proceedings of the BabyLM Challenge at the 27th Conference on Computational Natural Language Learning},
  pages = {1--34},
  address = {Singapore},
  publisher = {Association for Computational Linguistics},
  year = {2023},
  url = {https://aclanthology.org/2023.conll-babylm.1/}
}

@inproceedings{yam2024what,
  title = {What Should Baby Models Read? Exploring Sample-Efficient Data Composition on Model Performance},
  author = {Yam, Hong Meng and Paek, Nathan},
  booktitle = {The 2nd BabyLM Challenge at the 28th Conference on Computational Natural Language Learning},
  pages = {284--291},
  address = {Miami, FL, USA},
  publisher = {Association for Computational Linguistics},
  year = {2024},
  url = {https://aclanthology.org/2024.conll-babylm.25/}
}

@inproceedings{charpentier2024gptbert,
  title = {{GPT} or {BERT}: Why Not Both?},
  author = {Charpentier, Lucas Georges Gabriel and Samuel, David},
  booktitle = {The 2nd BabyLM Challenge at the 28th Conference on Computational Natural Language Learning},
  pages = {262--283},
  address = {Miami, FL, USA},
  publisher = {Association for Computational Linguistics},
  year = {2024},
  url = {https://aclanthology.org/2024.conll-babylm.24/}
}

@inproceedings{dehghani2019universal,
  title = {Universal Transformers},
  author = {Dehghani, Mostafa and Gouws, Stephan and Vinyals, Oriol and Uszkoreit, Jakob and Kaiser, Lukasz},
  booktitle = {International Conference on Learning Representations},
  year = {2019},
  url = {https://openreview.net/forum?id=HyzdRiR9Y7}
}

@inproceedings{lan2020albert,
  title = {{ALBERT}: A Lite {BERT} for Self-Supervised Learning of Language Representations},
  author = {Lan, Zhenzhong and Chen, Mingda and Goodman, Sebastian and Gimpel, Kevin and Sharma, Piyush and Soricut, Radu},
  booktitle = {International Conference on Learning Representations},
  year = {2020},
  url = {https://openreview.net/forum?id=H1eA7AEtvS}
}

@inproceedings{yang2024looped,
  title = {Looped Transformers are Better at Learning Learning Algorithms},
  author = {Yang, Liu and Lee, Kangwook and Nowak, Robert and Papailiopoulos, Dimitris},
  booktitle = {International Conference on Learning Representations},
  year = {2024},
  url = {https://openreview.net/forum?id=XpVoUnPuYV}
}

@article{kohli2026loop,
  title = {Loop, Think, \& Generalize: Implicit Reasoning in Recurrent-Depth Transformers},
  author = {Kohli, Harsh and Parthasarathy, Srinivasan and Sun, Huan and Yao, Yuekun},
  journal = {arXiv preprint arXiv:2604.07822},
  year = {2026},
  doi = {10.48550/arXiv.2604.07822},
  url = {https://arxiv.org/abs/2604.07822}
}

@article{warstadt2020blimp,
  title = {{BLiMP}: The Benchmark of Linguistic Minimal Pairs for {E}nglish},
  author = {Warstadt, Alex and Parrish, Alicia and Liu, Haokun and Mohananey, Anhad and Peng, Wei and Wang, Sheng-Fu and Bowman, Samuel R.},
  journal = {Transactions of the Association for Computational Linguistics},
  volume = {8},
  pages = {377--392},
  year = {2020},
  doi = {10.1162/tacl_a_00321},
  url = {https://aclanthology.org/2020.tacl-1.25/}
}

@article{ivanova2024ewok,
  title = {Elements of World Knowledge ({EWoK}): A Cognition-Inspired Framework for Evaluating Basic World Knowledge in Language Models},
  author = {Ivanova, Anna A. and Sathe, Aalok and Lipkin, Benjamin and Kumar, Unnathi and Radkani, Setayesh and Clark, Thomas H. and Kauf, Carina and Hu, Jennifer and Pramod, R. T. and Grand, Gabriel and others},
  journal = {arXiv preprint arXiv:2405.09605},
  year = {2024},
  url = {https://arxiv.org/abs/2405.09605}
}

@inproceedings{kim2023entity,
  title = {Entity Tracking in Language Models},
  author = {Kim, Najoung and Schuster, Sebastian},
  booktitle = {Proceedings of the 61st Annual Meeting of the Association for Computational Linguistics (Volume 1: Long Papers)},
  pages = {3835--3855},
  address = {Toronto, Canada},
  publisher = {Association for Computational Linguistics},
  year = {2023},
  doi = {10.18653/v1/2023.acl-long.213},
  url = {https://aclanthology.org/2023.acl-long.213/}
}

@inproceedings{wang2019glue,
  title = {{GLUE}: A Multi-Task Benchmark and Analysis Platform for Natural Language Understanding},
  author = {Wang, Alex and Singh, Amanpreet and Michael, Julian and Hill, Felix and Levy, Omer and Bowman, Samuel R.},
  booktitle = {International Conference on Learning Representations},
  year = {2019},
  url = {https://openreview.net/forum?id=rJ4km2R5t7}
}

@inproceedings{you2020large,
  title = {Large Batch Optimization for Deep Learning: Training {BERT} in 76 Minutes},
  author = {You, Yang and Li, Jing and Hseu, Jonathan and Song, Xiaodan and Demmel, James and Hsieh, Cho-Jui},
  booktitle = {International Conference on Learning Representations},
  year = {2020},
  url = {https://openreview.net/forum?id=Syx4wnEtvH}
}

@inproceedings{yu-etal-2024-antlm,
  title = {{A}nt{LM}: Bridging Causal and Masked Language Models},
  author = {Yu, Xinru and Guo, Bin and Luo, Shiwei and Wang, Jie and Ji, Tao and Wu, Yuanbin},
  booktitle = {The 2nd BabyLM Challenge at the 28th Conference on Computational Natural Language Learning},
  pages = {324--331},
  address = {Miami, FL, USA},
  publisher = {Association for Computational Linguistics},
  year = {2024},
  url = {https://aclanthology.org/2024.conll-babylm.29/}
}

@inproceedings{hu-etal-2024-findings,
  title = {Findings of the Second {B}aby{LM} Challenge: Sample-Efficient Pretraining on Developmentally Plausible Corpora},
  author = {Hu, Michael Y. and Mueller, Aaron and Ross, Candace and Williams, Adina and Linzen, Tal and Zhuang, Chengxu and Choshen, Leshem and Cotterell, Ryan and Warstadt, Alex and Wilcox, Ethan Gotlieb},
  booktitle = {The 2nd BabyLM Challenge at the 28th Conference on Computational Natural Language Learning},
  pages = {1--21},
  address = {Miami, FL, USA},
  publisher = {Association for Computational Linguistics},
  year = {2024},
  url = {https://aclanthology.org/2024.conll-babylm.1/}
}

@inproceedings{haga-etal-2024-babylm,
  title = {{B}aby{LM} Challenge: Exploring the Effect of Variation Sets on Language Model Training Efficiency},
  author = {Haga, Akari and Fukatsu, Akiyo and Oba, Miyu and Bisazza, Arianna and Oseki, Yohei},
  booktitle = {The 2nd BabyLM Challenge at the 28th Conference on Computational Natural Language Learning},
  pages = {252--261},
  address = {Miami, FL, USA},
  publisher = {Association for Computational Linguistics},
  year = {2024},
  url = {https://aclanthology.org/2024.conll-babylm.23/}
}

@inproceedings{nair-etal-2024-babylm,
  title = {{B}aby{LM} Challenge: Experimenting with Self-Distillation and Reverse-Distillation for Language Model Pre-Training on Constrained Datasets},
  author = {Nair, Aakarsh and Hancharova, Alina and Kumar, Mayank and Gharaee, Ali},
  booktitle = {The 2nd BabyLM Challenge at the 28th Conference on Computational Natural Language Learning},
  pages = {28--36},
  address = {Miami, FL, USA},
  publisher = {Association for Computational Linguistics},
  year = {2024},
  url = {https://aclanthology.org/2024.conll-babylm.3/}
}

@inproceedings{ghanizadeh-dousti-2024-towards,
  title = {Towards Data-Efficient Language Models: A Child-Inspired Approach to Language Learning},
  author = {Ghanizadeh, Mohammad Amin and Dousti, Mohammad Javad},
  booktitle = {The 2nd BabyLM Challenge at the 28th Conference on Computational Natural Language Learning},
  pages = {22--27},
  address = {Miami, FL, USA},
  publisher = {Association for Computational Linguistics},
  year = {2024},
  url = {https://aclanthology.org/2024.conll-babylm.2/}
}

@inproceedings{edman-etal-2024-babylms,
  title = {Are {B}aby{LM}s Second Language Learners?},
  author = {Edman, Lukas and Bylinina, Lisa and Ghorbanpour, Faeze and Fraser, Alexander},
  booktitle = {The 2nd BabyLM Challenge at the 28th Conference on Computational Natural Language Learning},
  pages = {166--173},
  address = {Miami, FL, USA},
  publisher = {Association for Computational Linguistics},
  year = {2024},
  url = {https://aclanthology.org/2024.conll-babylm.14/}
}

\end{document}